\documentclass[letterpaper, 10 pt, conference]{ieeeconf}  
\IEEEoverridecommandlockouts                              
\usepackage[utf8]{inputenc}
\usepackage[T1]{fontenc}
\usepackage{float}
\usepackage{lipsum}
\usepackage{graphicx}
\usepackage{booktabs}
\usepackage{subcaption}
\usepackage{makecell}
\usepackage{graphics} 
\usepackage{epsfig} 
\usepackage{mathptmx} 
\usepackage{mathptmx} 
\usepackage{amsmath} 
\usepackage{amssymb}  
\usepackage{xspace}

\makeatletter

\title{\LARGE \bf
Panoptic Lintention Network: Towards Efficient Navigational Perception for the Visually Impaired
}

\author{Wei Mao$^{*1,2}$, Jiaming Zhang$^{*1}$, Kailun Yang$^{1}$ and Rainer Stiefelhagen$^{1}$
\thanks{This work was supported in part by the Federal Ministry of Labor and Social Affairs (BMAS) through the AccessibleMaps project under Grant 01KM151112, in part by the University of Excellence through the ``KIT Future Fields'' project, in part by the Hangzhou SurImage Technology Company Ltd., and in part by the Hangzhou KrVision Technology Company Ltd. (krvision.cn).}
\thanks{* denotes equal contribution. $^{1}$The authors are with Institute for Anthropomatics and Robotics, Karlsruhe Institute of Technology, 76131 Karlsruhe, Germany; $^{2}$The author is also with Sino-German School, Tongji University, Shanghai 201800, China (correspondence: kailun.yang@kit.edu).}
}

\begin{document}

\maketitle
\thispagestyle{empty}
\pagestyle{empty}

\begin{abstract}
Classic computer vision algorithms, instance segmentation, and semantic segmentation can not provide a holistic understanding of the surroundings for the visually impaired. In this paper, we utilize panoptic segmentation to assist the navigation of visually impaired people by offering both things and stuff awareness in the proximity of the visually impaired efficiently. To this end, we propose an efficient Attention module -- Lintention which can model long-range interactions in linear time using linear space. Based on Lintention, we then devise a novel panoptic segmentation model which we term Panoptic Lintention Net. Experiments on the COCO dataset indicate that the Panoptic Lintention Net raises the Panoptic Quality (PQ) from 39.39 to 41.42 with 4.6\% performance gain while only requiring 10\% fewer GFLOPs and 25\% fewer parameters in the semantic branch. Furthermore, a real-world test via our designed compact wearable panoptic segmentation system, indicates that our system based on the Panoptic Lintention Net accomplishes a relatively stable and exceptionally remarkable panoptic segmentation in real-world scenes.
\end{abstract}

\section{Introduction}
According to the world health organization, at least 2.2 billion people have a vision impairment or
blindness~\cite{bourne2017magnitude}.
The majority of people with visual impairments still use simple and conventional assistive tools e.g., white canes which cannot provide any semantic interpretation of their surroundings.
Therefore, it is quite meaningful to supply the visually impaired with high-level scene understanding for their safety in daily life, especially in traffic scenarios \cite{yang2018unifying}.   

\begin{figure}[t!]
    \centering
    \begin{subfigure}[b]{0.99\columnwidth}   
		\centering 
	    \includegraphics[width=\columnwidth]{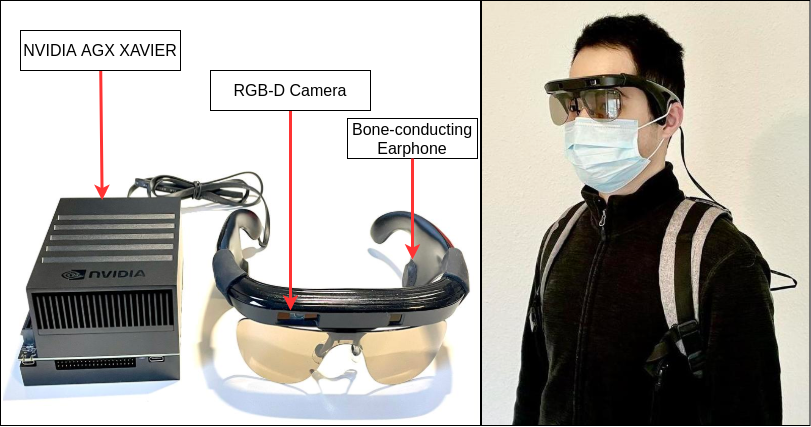}
	    \vskip-1ex
		\caption[]{\small Wearable system}    
		\label{fig:system}
	\end{subfigure}
    \begin{subfigure}[b]{0.8\columnwidth}   
		\includegraphics[width=\columnwidth]{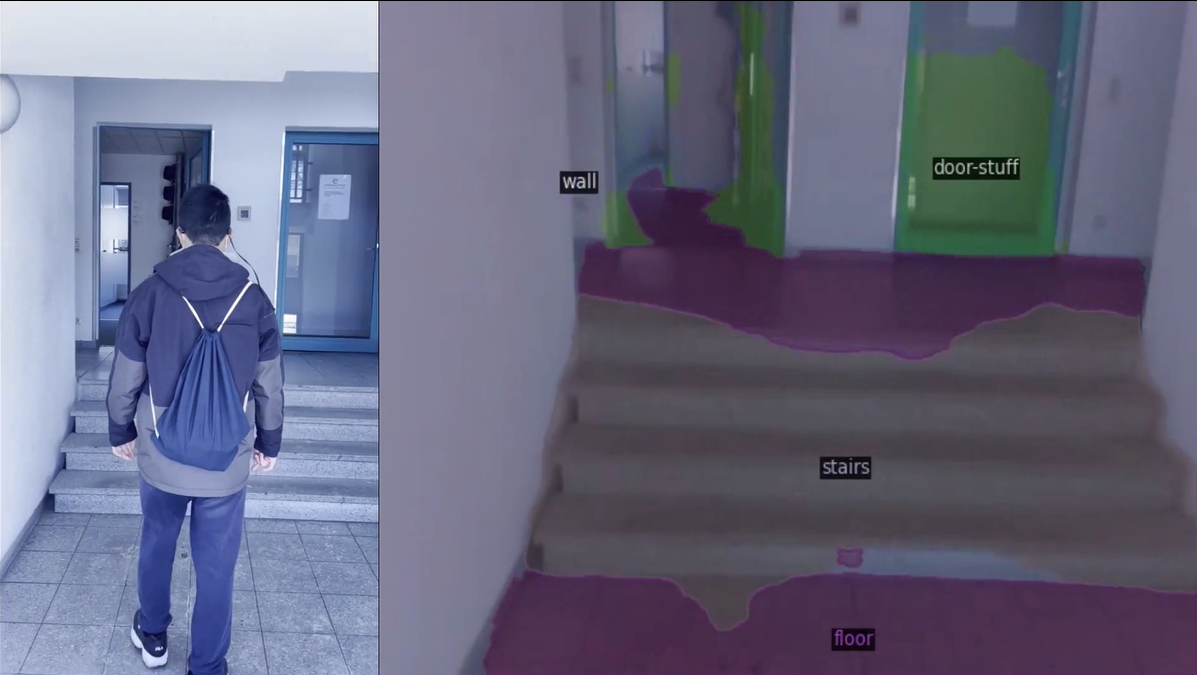}
		\vskip-1ex
		\caption[]{\small Indoor scene}    
		\label{fig:indoor}
	\end{subfigure}
	\begin{subfigure}[b]{0.8\columnwidth}   
		\centering 
		\includegraphics[width=\columnwidth]{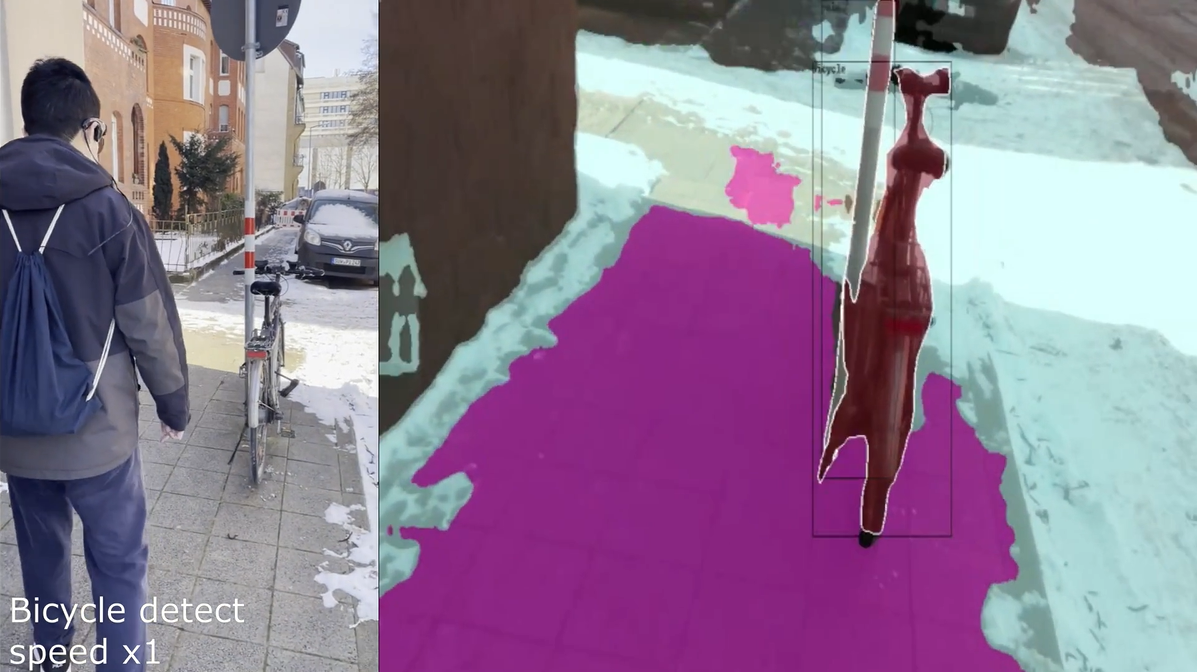}
		\vskip-1ex
		\caption[]{\small Outdoor scene}    
		\label{fig:outdoor}
	\end{subfigure}
	\vskip-1ex
    \caption{Illustration of (a) our wearable assistive system for visually impaired people. (b) shows the segmentation of stair, floor, door, and various objects for indoor navigation, while (c) shows the segmentation of walkable sidewalk and bicycle in a snowy outdoor street navigation scenario. }
    \label{fig:wearable_system}
\vskip-4ex
\end{figure}

To date, navigational perception for the visually impaired is still primarily founded on computer vision techniques, either classic computer vision algorithms or modern deep learning-based ones. The early approaches~\cite{rodriguez2012assisting,schwarze2015intuitive} mainly attached great importance to the so-called generic object detection where the generic objects range from foreground obstacles to traversable space without considering the semantics for concrete segments~\cite{watson2020footprints} or with  a limited level of the semantics such as pedestrian detection ~\cite{lee2020pedestrian}.
In the real world, visually impaired people constantly encounter various sorts of objects, especially in outdoor environments. 
To inform them of what exact sorts of objects they are facing, the fast growing deep learning-based computer vision techniques can be exploited.
For instance, the popular instance segmentation framework Mask R-CNN~\cite{he2017mask} has been adopted to aid visually impaired people to identify a wide spectrum of objects which may occur in their surroundings~\cite{long2019unifying}.
Apart from the countable objects like pedestrians, vehicles and such, collectively termed things, the stuff, amorphous regions of similar texture or material such as road, sky, and grass~\cite{kirillov2019panoptic}, should also be incorporated into the visual assistance task on the grounds that the identification of traffic infrastructures like road and sidewalk is essential for the visually impaired to participate in traffic safely.
As an example of stuff perception for blind people, a light-weight semantic segmentation network was proposed to simultaneously address the detection of blind roads and crosswalks specifically~\cite{cao2020rapid}, but overlooking the thing classes like various obstacles in the scene.

Naturally, it leads us to the question: \textit{how to unify the recognition tasks for the typically distinct thing classes and stuff classes in the neighborhood of the visually impaired people?} Fortunately,
the panoptic segmentation
unifies instance and semantic segmentation problems by mapping each pixel in a image to a pair of class label and instance ID~\cite{kirillov2019panoptic}, which serve as comprehensive visual perception information for navigating the visually impaired.
In this paper, we propose an efficient panoptic segmentation model called Panoptic Lintention Net and verify its performance via our designed wearable assistive system in real-world navigation scenarios (see Fig.~\ref{fig:wearable_system}).

\section{Related Work}

\subsection{Visual Assistance}

In early assistance systems for visually impaired people via scene understanding, the main goal is to notify the traversable directions.
Rodriguez et al.~\cite{rodriguez2012assisting} segmented the image into background and obstacles based on dense disparity maps and ground plane estimation algorithms.
Schwarze et al.~\cite{schwarze2015intuitive} fulfilled the perception of the surroundings by combining various complementary methods such as estimation and tracking of the geometric scene background, generic object detection and ego-pose estimation, collectively offering more information about the scene to the visually impaired.
Nonetheless, these approaches based on classical vision algorithms do not distinguish specific instances such as pedestrians, cars, bikes, etc.

Recent assistance approaches for the visually impaired emerge with the renaissance of deep learning~\cite{duh2020v}.
Yohannes et al.~\cite{yohannes2019content} applied the notable instance segmentation framework Mask R-CNN~\cite{he2017mask} to recognize various objects in the outdoor environment.
Yet, instance segmentation internally neglects stuff classes including road and sidewalk, which are safety-critical objects for the walking visually impaired outdoors. 
In addition to instance segmentation, recent years witness the wide use of semantic segmentation due to the thriving deep learning.
Yang et al.~\cite{yang2018intersection} leveraged real-time semantic segmentation to coordinate the perception needs of visually impaired pedestrians at traffic intersections.
Additionally, Cao et al.~\cite{cao2020rapid} developed a light-weight network for fast detection of blind road and sidewalk. Nevertheless, semantic segmentation provides incomplete visual perception by not making any distinction between multiple instances of the same category inherently.

\subsection{Panoptic Segmentation}

Benefiting from the advances in both semantic and instance segmentation, the proposer of the panopitc segmentation task used a naive solution, namely training two separate leading models for semantic and instance segmentation independently and then fusing the semantic masks and instance masks to obtain the final panoptic mask~\cite{kirillov2019panoptic}.
Such solution is conceptually simple and easy to implement but doubles the computational cost and the model complexity, which leads to the effort to devise a joint solution.

To date, many panoptic segmentation methods are proposed based on the well-established instance segmentation models like Mask R-CNN~\cite{he2017mask}.
Panoptic FPN~\cite{kirillov2019pfpn} is proposed to perform the panoptic segmentation by attaching an additional semantic segmentation branch to the original Mask R-CNN.
As Mask R-CNN has become a strong baseline for instance segmentation, Panoptic FPN is also deemed to be a strong baseline for the panoptic segmentation task.
Numerous subsequent panoptic segmentation methods are designed to enhance Panoptic FPN with additional techniques such as AUNet~\cite{li2019attention} and BANet~\cite{chen2020banet}.

Aside from constructing a panoptic segmentation model based on instance segmentation, it is also quite natural to build a panoptic segmentation architecture on top of existing semantic segmentation models.
Panoptic-DeepLab \cite{cheng2020panoptic} is a typical example of developing a panoptic segmentation model from a semantic segmentation architecture,  which adopts the dual-ASPP and dual-decoder structures specific to semantic- and instance segmentation, respectively.

\section{Panoptic Lintention Network}

\subsection{Linear Attention}
 The Panoptic FPN~\cite{kirillov2019panoptic} only exploits the standard convolution to construct the semantic segmentation branch. 
 To enhance the performance, we attempt to design an novel attention module to model long-range interactions like multi-scale features or global information, with less compute time and less space instead of the quadratic complexity in the standard Self-Attention~\cite{vaswani2017attention}.

\newcommand{\RNum}[1]{\uppercase\expandafter{\romannumeral #1\relax}}
\newcommand{\R}[1]{\mathbb{R}^{#1}}
 
In the real world, the number of stuff and thing classes is limited, and not all pixels in one instance segment or stuff segment are necessary for us to identify the thing or stuff category.
Thus, we assume that a feature map can be described as a set of semantic groups which are essential for visual recognition and the count of semantic groups is significantly smaller than the total number of pixels.
Specifically, we set the number of semantic groups to be 16 in this paper.
However, the larger number of semantic groups is expected to give more representational power but need more compute and memory resources.
Under this assumption, we can first transform the input feature map into a set of semantic groups and then compare each pixel to these semantic groups by computing the scaled dot products (attention scores) \cite{vaswani2017attention}, which is relatively efficient in time and space thanks to the limited number of semantic groups.
Taking advantage of these attention scores, 
we adopt similar procedures in the standard Self-Attention to yield the output with the same resolution as the input feature map.
As proved in the following, our proposed attention module enjoys linear complexity in both time and space.
Hence, we name our proposed attention \emph{Linear Attention}, \emph{Lintention} for short, which is split into 
into five sub-modules:
Query Generator (GQ), Key Generator (KG), Value Generator (VG), Attention Scores Generator (ASG), and Result Generator (RG).
Given the input feature maps $F \in \mathbb{R}^{N \times C \times H \times W}$, where $N$ is the mini-batch size, $C$ is the channel size, $H$ and $W$ specify the height
and width of the feature map, 
a dimension permutation on it results in a new feature map $X \in \R{N \times H \times W \times C}$, which is fed into the following Lintention sub-modules.

\textbf{Query Generator (QG)}
The input  $X \in \R{N \times H \times W \times C}$ is linearly projected with a learned weight matrix into queries $Q \in \R{N \times H \times W \times C}$, which is given by 
\begin{equation}
Q = X *_{(nhwc, cd, nhwd)} W^Q
\label{eqn:Q_gen}
\end{equation}
where $W^Q \in \R{C \times C}$ is the learned query projection matrix, and $*_{(\cdot, \cdot, \cdot)}$ denotes the extended  Einstein Notation \cite{laue2020simple}. 

As with other deep learning models, we are also concerned with the time complexity, space complexity, and model complexity, which are denoted by 
$T(\cdot)$, $S(\cdot)$ and $M(\cdot)$ in this paper.
After inspecting Equ.~\ref{eqn:Q_gen}, we can obtain the complexity of the Query Generator as follows:
\begin{IEEEeqnarray}{rCl}
T(\text{QG}) & = & 2NHWC^2 = \mathcal{O}(NHWC^2) \label{eqn:Q_T}\\
S(\text{QG}) & = & NHWC = \mathcal{O}(NHWC) \label{eqn:Q_S}\\
M(\text{QG}) & = & C^2 \label{eqn:Q_M}
\end{IEEEeqnarray}

\textbf{Key Generator (KG)}
Instead of  naive subsampling by interpolation, we first apply a linear classifier to classify each pixel in the feature map, and then aggregate information along the spatial dimension from both queries $Q$ and classification results $L$ into a set of semantic groups.

The per-pixel classification for the input feature maps $X \in \R{N \times H \times W \times C}$ is given by 
\begin{equation}
    L = \text{softmax}(X *_{(nhwc, cp, nhwp)}W^K)
    \label{eqn:sem_gp_gen}
\end{equation}
where $L \in \R{N \times H \times W \times P}$ is the classification result, where $L(n,h,w, :)$
is interpreted as the class probability distribution for $(h, w)$ pixel in $n$th input of the batch; $W^K \in \R{C \times P}$ is the linear classifier, which classifies each pixel as one of the $P$ semantic groups; softmax is used to normalize the classification scores into the range of 0 to 1 along the last dimension of  $L$; 

The $p$th semantic group in the $n$th input of the batch
is now expressed by $L(n, :, :, p)$, which is rank-2 tensor across the whole spatial dimension.
In order to obtain a compact representation for each semantic group,
we contract the spatial dimension by absorbing the information from the queries deriving from the input feature maps and have 
\begin{equation}
    K = L *_{(nhwp, nhwc, npc)} Q 
    \label{eqn:key_gen}
\end{equation}
where $K \in \R{N \times P \times C}$ is the key tensor in our proposed Lintention; 
$L \in \R{N \times H \times W \times P}$ is the classification result in Equ.~\ref{eqn:sem_gp_gen}; 
$Q \in \R{N \times H \times W \times C}$ is the query tensor generated by Equ.~\ref{eqn:Q_gen}.

Our Lintention generates only $P$ keys, each of which corresponds to one learned semantic group.
Note that each value of $K$ is related to all pixels in the feature map, since we sum over the whole spatial dimensions.

The overall complexity of 
Key Generator KG is as follows:
\begin{IEEEeqnarray}{rCl}
T(\text{KG}) & = & 4NHWCP + 2NHWP= \mathcal{O}(NHWCP) \label{eqn:KGI_T} \\
S(\text{KG}) & = & NHWP + NPC = \mathcal{O}(NP(HW+C)) \label{eqn:KGI_S}\\
M(\text{KG}) & = & CP \label{eqn:KGI_M}
\end{IEEEeqnarray}
where the second term of the Equ.~\ref{eqn:KGI_T} is caused by the softmax.

\textbf{Value Generator (VG)}
Considering that we must have the same number of values as that of the keys,
values in our Lintention are obtained by linearly projecting the keys $K$ with a learned matrix, which is given by
\begin{equation}
    V = K *_{(npc, cd, npd)} W^V
    \label{eqn: Value_Gen}
\end{equation}
where $V \in \R{N \times P \times C}$ is the learned values from the keys $K$ by a value projection matrix $W^V \in \R{C \times C}$.
Similar to the standard Self-Attention, the final output vector is the weighted sum (linear combination) of these values. 
Note that each value of $V$ is also related to all pixels in the feature map as each value of $K$ aggregates information from all pixels. 

The complexities of VG can also be figured out as follows:
\begin{IEEEeqnarray}{rCl}
T(\text{VG}) & = & 2NPC^2 = \mathcal{O}(NPC^2) \label{eqn:V_T}\\
S(\text{VG}) & = & NPC = \mathcal{O}(NPC) \label{eqn:V_S}\\
M(\text{VG}) & = & C^2 \label{eqn:V_M}
\end{IEEEeqnarray}

\textbf{Attention Score Generator (ASG)}
The attention matrix, or rather the softmax, is the root of the quadratic complexity in both time and space.
To counter this issue, we have reduced the number of keys from $HW$ to $P$, where $HW$ is the count of pixels and $P$ is the number of semantic groups.
Hence, each pixel of the input only needs to attend to a much smaller number of learned semantic groups.
Regarding the measure for the similarity between a query and a key,
we follow the scaled dot product similarity used in the standard Self-Attention module \cite{vaswani2017attention}.
Specifically, our attention scores are given by
\begin{equation}
    AttScores = \text{softmax}(\frac{K *_{(npc, nhwc, nhwp)}Q}{\sqrt{C}})
    \label{eqn:Linatt_att_scores}
\end{equation}
where $AttScores \in \R{N \times H \times W \times P}$ is the tensor of attention scores, where $AttScores(n, h, w, p)$ measures the similarity between $(h,  w)$ pixel of the $n$th input feature map of the batch and $p$th learned semantic group; 
$K$ and $Q$ are the keys and queries;
$C$ is the number of channels of the input feature map.
We see each pixel as a word embedded into a $C$-dimensional vector in the latent space.
The scaling factor $\frac{1}{\sqrt{C}}$ is introduced for the purpose of pre-normalizing the scores and boosting the gradient flow through the softmax.

We count each exponentiation of softmax as one FLOP.
As a result, the complexities of ASG is given by 
\begin{IEEEeqnarray}{rCl}
T(\text{ASG}) & = & 2NHWPC + 3NHWP = \mathcal{O}(NHWPC) \label{eqn:ASG_T}\\
S(\text{ASG}) & = & NHWP = \mathcal{O}(NHWP) \label{eqn:ASG_S}\\
M(\text{ASG}) & = & 0 \label{eqn:ASG_M}
\end{IEEEeqnarray}
where term $3NHWP$ in Equ.~\ref{eqn:ASG_T} accounts for the scaling by $\frac{1}{\sqrt{C}}$ and softmax.

\textbf{Result Generator (RG)}
After working out the attention scores between each query and each key,
we exploit the strategy in the standard Self-Attention to get the final results of our Lintention, as given by 
\begin{equation}
    Y = \text{Lintention}(X) = AttScores *_{(nhwp, npc, nhwc)} V  
    \label{eqn: Lintention}
\end{equation}
where $Y \in \R{N \times H \times W \times C}$ is the output of the Lintention module; $AttScores$ denotes the attention scores generated by ASG and $V$ represents the
values generated by VG.

\begin{IEEEeqnarray}{rCl}
T(\text{RG}) & = & 2NHWPC = \mathcal{O}(NHWPC) \label{eqn:RG_T}\\
S(\text{RG}) & = & NHWC = \mathcal{O}(NHWC) \label{eqn:RG_S}\\
M(\text{RG}) & = & 0 \label{eqn:RG_M}
\end{IEEEeqnarray}

Summing up the complexities of all above 5 Generators, we obtain the time, space, and model complexities of our proposed Lintention as follows:
\begin{IEEEeqnarray}{rCl}
T(\text{L}) & = & NHW(2C^2 + 8PC + 5P) + 2NPC^2 = \mathcal{O}(HW) \label{eqn:LinATT_T}\\
S(\text{L}) & = & 2NHWP + 2NHWC + 2NPC = \mathcal{O}(HW) \label{eqn:LinATT_S}\\
M(\text{L}) & = &  2C^2 + CP\label{eqn:LinATT_M}
\end{IEEEeqnarray}
where we only view the total number of pixels $HW$ as the input size of our Lintention.

\begin{figure}[t]
    \centering
    \includegraphics[width=\linewidth]{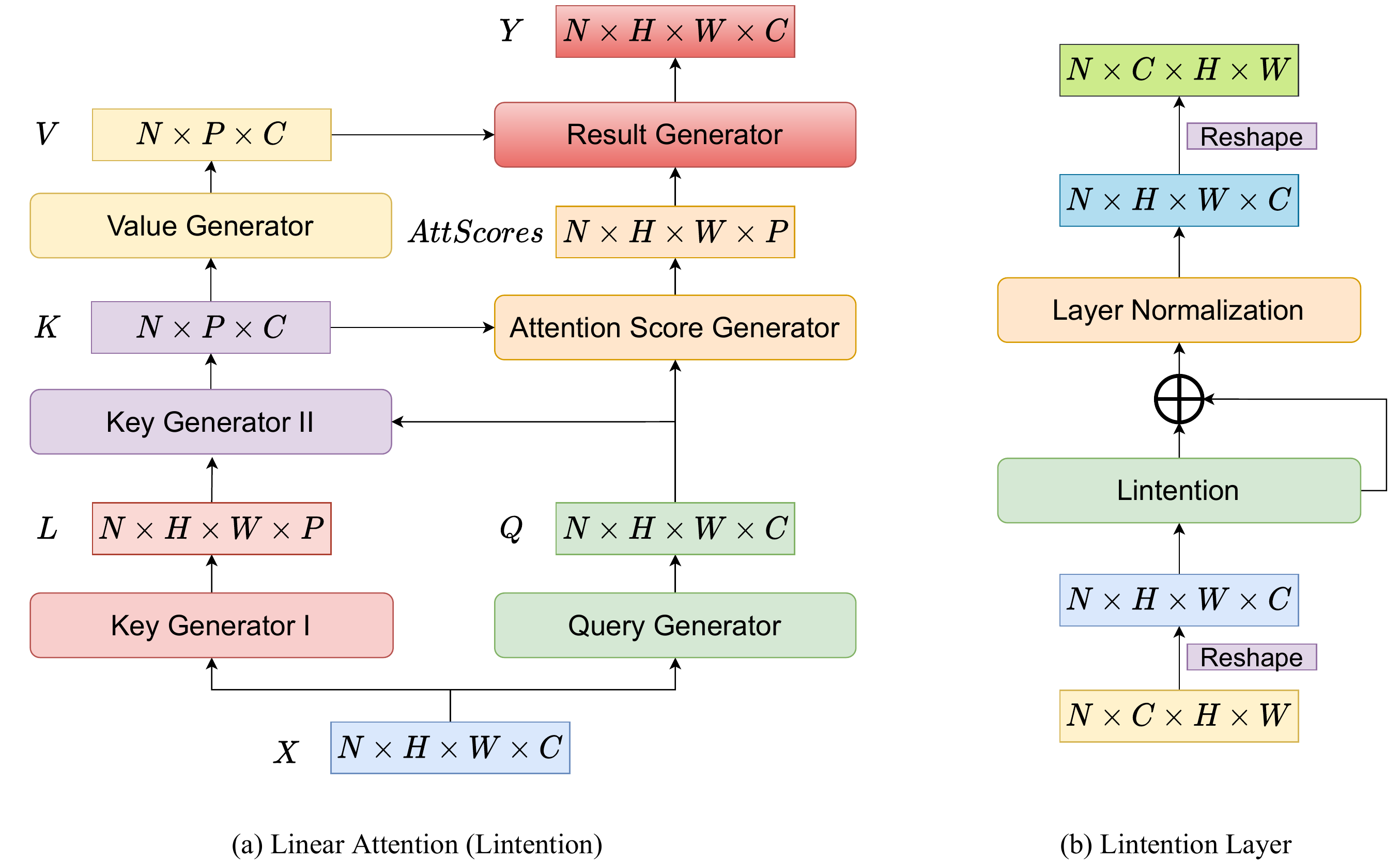}
    \vskip-1ex
    \caption{Lintention and Lintention Layer.}
    \label{fig:lintention_layer}
    \vskip-4ex
\end{figure}
To sum up, our proposed attention module, which we call Linear Attention, shortened to Lintention, is capable of modeling long-range interaction in linear time using linear memory with respect to the total number of pixels.
The overall architecture of our Lintention is demonstrated in Fig.~\ref{fig:lintention_layer}(a).

\subsection{Panoptic Lintention Net}
To apply our Lintention Module to the panoptic segmentation network,
we design a Lintention Layer as illustrated in Fig.~\ref{fig:lintention_layer}(b).
As with the residual connection in ResNet,
we also add a skip connection to our Lintention, which boosts the gradient flow during training.
Furthermore, we apply Layer Normalization to modulate the output from the Lintention.
To be compatible with the shape of some CNN layers, we also perform reshape operation if necessary.

\begin{figure}[t]
    \centering
    \includegraphics[width=\linewidth]{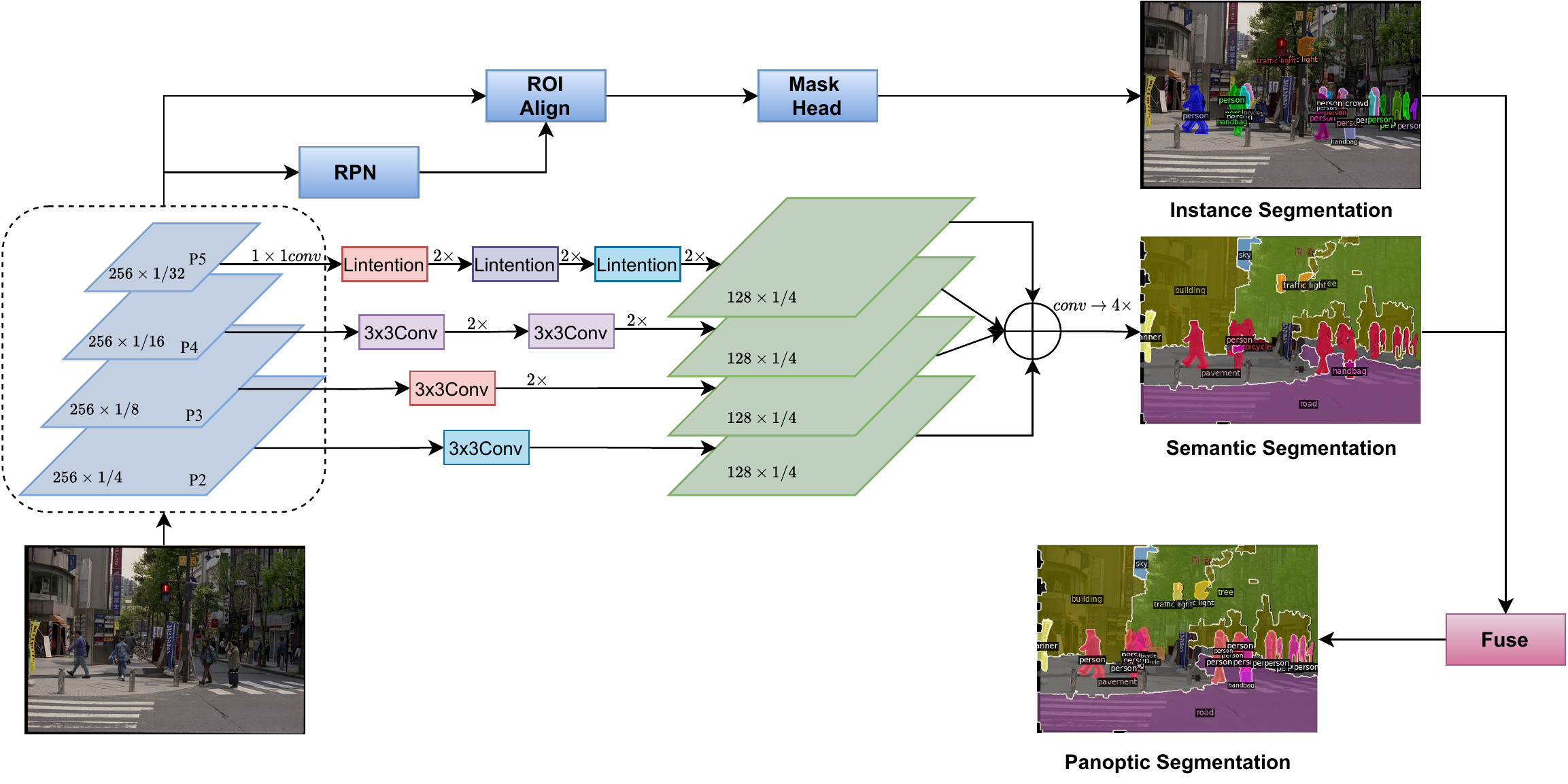}
    \vskip-1ex
    \caption{Panoptic Lintention Network.}
    \label{fig:pan_lin_net}
    \vskip-4ex
\end{figure}

We use ResNet-FPN as backbone to extract multi-scale feature maps $\{P_2, P_3, P_4, P_5\}$, whose scales are 1/4, 1/8, 1/16, 1/32 of the input image resolution, respectively.
Feature map at scale 1/32 loses small details but captures high-level semantics, which lends itself more to modeling long-range interaction.
Thus, we decide to attach Lintention Layer to
feature map at 1/32 scale, as indicated in the Fig.~\ref{fig:pan_lin_net}.
Through such configuration, we utilize the advantages of both Lintention and high-level FPN feature maps to model long-range dependencies.
Specifically, we first apply a $1\times 1$ convolution to reduce the input channel size from 256 to 128.
Then, we perform three upsampling stages to yield
a feature map at 1/4 scale, where each upsampling stage
consists of a Lintention layer and 2× bilinear upsampling. 
Each Lintention layer preserves the spatial size and capture long-range dependencies with linear complexity.
For FPN scales 1/16, 1/8, and 1/4, we adopt the same procedure as that in the Mask R-CNN \cite{he2017mask}.
We term the novel architecture for panoptic segmentation Panoptic Lintention Network as depicted in Fig.~\ref{fig:pan_lin_net}.

\section{Experiments}
\subsection{Various Models}
\begin{table*}[t]
    \centering
    \caption{Comparisons of Panoptic Segmentation Performance on the COCO~\cite{lin2014microsoft} {\em val} set.}
    \vskip-1ex
    \label{tab:coco_val}
    \resizebox{\textwidth}{!}{%
    \begin{tabular}{lc|ccc|ccc|ccc}
    \toprule
    Method & Backbone & PQ & SQ & RQ & PQ$^\mathrm{th}$ & SQ$^\mathrm{th}$ & RQ$^\mathrm{th}$ & PQ$^\mathrm{st}$ & SQ$^\mathrm{st}$ & RQ$^\mathrm{st}$ \\
    \midrule
   Panoptic FPN~\cite{kirillov2019pfpn}  & Res50-FPN & 39.39 & 77.84 & 48.31 & 45.91 & 80.86 & 55.34 & 29.55 & 73.27 & 37.69 \\
   Panoptic PyConv       & Res50-FPN & 40.49 & 78.41 & 49.45 & 46.87 & 81.12 & 56.31 & 30.86 & \textbf{74.32} & 39.09 \\
   Panoptic VerConv      & Res50-FPN & 40.54 & 78.44 & 49.49 & 47.07 & 81.19 & 56.51 & 30.69 & 74.29 & 38.88 \\
   Panoptic VerConvSep   & Res50-FPN & 40.51 & 78.43 & 49.46 & 47.17 & 81.18 & 56.69 & 30.47 & 74.27 & 38.54 \\
   Panoptic Lintention   & Res50-FPN & \textbf{41.42} & \textbf{78.55} & \textbf{50.52} & \textbf{48.39} & \textbf{82.23} & \textbf{58.04} & \textbf{30.89} & 72.98 & \textbf{39.17} \\
     \bottomrule
    \end{tabular}%
    }
    \vskip-3ex
\end{table*}
Our fundamental idea is to augment the well-established instance segmentation model Mask R-CNN~\cite{he2017mask} with an efficient semantic segmentation branch.
The standard convolution is internally unable to process multi-scale details due to its single type of kernel with a single spatial extent and depth, which is the main obstacle to a better-performing semantic segmentation branch.
In contrast, the Pyramidal Convolution (PyConv)~\cite{duta2020pyramidal} is designed to capture multi-scale details in the feature map via a pyramid of grouped convolutions with various kernel sizes and depths without increasing the time and model complexity.
Given that the reliance of PyConv on the grouped convolution hampers the channel interactions among different groups,
we further design a novel convolution operator termed Versatile Convolution (VerConv) to refine the PyConv by explicitly modeling the channel interdependencies
by applying SE-like operation \cite{hu2018squeeze} to the outputs of PyConv collectively.
Additionally, we also design a variant of VerConv -- VerConvSep where the SE-like operation is applied to output of each PyConv level separately.
By substituting the standard convolution in the semantic branch of the Panoptic FPN \cite{kirillov2019pfpn} with PyConv, VerConv, VerConvSep, respectively,
we obtain three sorts of panoptic segmentation models: Panoptic PyConv Network, Panoptic VerConv Network, and Panoptic VerConvSep Network, in addition to the Panoptic Lintention Network.

\subsection{Implementation Details}
We initialize our backbone model with weights extracted from PyTorch's ImageNet-pretrained ResNet-50~\cite{he2016deep}.
Our designed models: Panoptic PyConv Net, Panoptic VerConv Net and Panoptic VerConvSep Net are trained by SGD optimizer 
with momentum of 0.9 using a fixed schedule of 270k iterations and base learning rate 0.01,
decreasing the learning rate by a factor of 10 after 180k and
240k iterations.  
At the beginning of training, 
we perform a warm-up phase where the learning rate is linearly increased from $\frac{0.01}{10^3}$ to 0.01 in 1000 iterations.
Given that model with attention mechanism takes extra-long training time to converge~\cite{carion2020end},
we double the training iterations for the Panoptic Lintention Net while adopting a similar learning rate schedule and warm-up strategy.
The weight decay is set to $1 \times 10^{-4}$ for adding an L2 penalty to the training loss.
Training of all the designed models are performed on batches of
4 images using a computing node equipped with 4 GTX1080Ti GPUs. 

\subsection{Comparative Analysis}
After training our proposed models on the COCO~\cite{lin2014microsoft} training set, we further run the trained models on the COCO validation set and compute the panoptic segmentation metric on all classes, thing and stuff classes, respectively.
Table~\ref{tab:coco_val} shows our main results on the COCO validation set.

From the table, we can see that all of the designed models outperform the baseline Panoptic FPN in terms of all kinds of PQ metrics, except that the Panoptic Lintentin performs worse on the SQ$^{st}$.
The PQ metric can be factorized into the product of SQ and RQ which measure segmentation and recognition quality, respectively~\cite{kirillov2019panoptic}.
 
Table~\ref{tab:coco_val} shows that our proposed Panoptic Lintention Net surpasses all the other four models in terms of all sorts of PQ metrics other than the SQ$^{\mathrm{st}}$ which measures the segmentation performance on the stuff class.
A predicted segment and a ground truth segment are defined to match only if their IoU is strictly greater than 0.5 \cite{kirillov2019panoptic}.
Although the segmentation performance on stuff classes, as measured by SQ$^{\mathrm{st}}$ for the Panoptic Lintention Net is even worse than the baseline by a small margin of 0.38, 
recognition performance on the stuff categories, as measured by RQ$^{\mathrm{st}}$ is significantly better than the baseline.
We can infer that it is relatively easier for our Panoptic Lintention Net to make the IoU of the predicted segment and ground truth segment on stuff classes greater than 0.5, which leads to a larger true positive
and in turn better RQ$^{\mathrm{st}}$ result, but hard to push the IoU to a high level, which in contrast yields lower SQ$^{\mathrm{st}}$ value.
we think the high RQ$^{\mathrm{st}}$ but low SQ$^{\mathrm{st}}$ is due to the better capability of the Lintention Module in learning long-range dependencies with some loss of fine-grained details.
\begin{table}[t]
    \centering
    \caption{Comparison of different semantic segmentation branches w.r.t. FLOPs and Parameters. The number of FLOPs is computed for a $960 \times 1280$ image. \dag ~ denotes the semantic segmentation branch in Panoptic FPN, namely our baseline. }
    \vskip-1ex
    \label{tab:flops_params_sem_head}
    \begin{tabular}{ccc}
    \toprule
    Method & (G)FLOPs & \# of Parameters \\
    \midrule
     P-FPN Sem. Branch$^{\dag}$ \cite{kirillov2019pfpn}  & 36.98  & 1.6M \\ 
     PyConv Sem. Branch       & 35.36   & 1.2M \\
     VerConv Sem. Branch       & 35.36  & 1.3M \\
     VerConvSep Sem. Branch    & 35.36  & 1.3M \\
     Lintention Sem. Branch     & \textbf{\makecell{33.28 (-10.0\%)}}  & \textbf{\makecell{1.2M  (-25.0\%)}} \\
    \bottomrule
    \end{tabular}
    \vskip-3ex
\end{table}

\begin{figure*}[t]
    \centering
    \includegraphics[width=\linewidth]{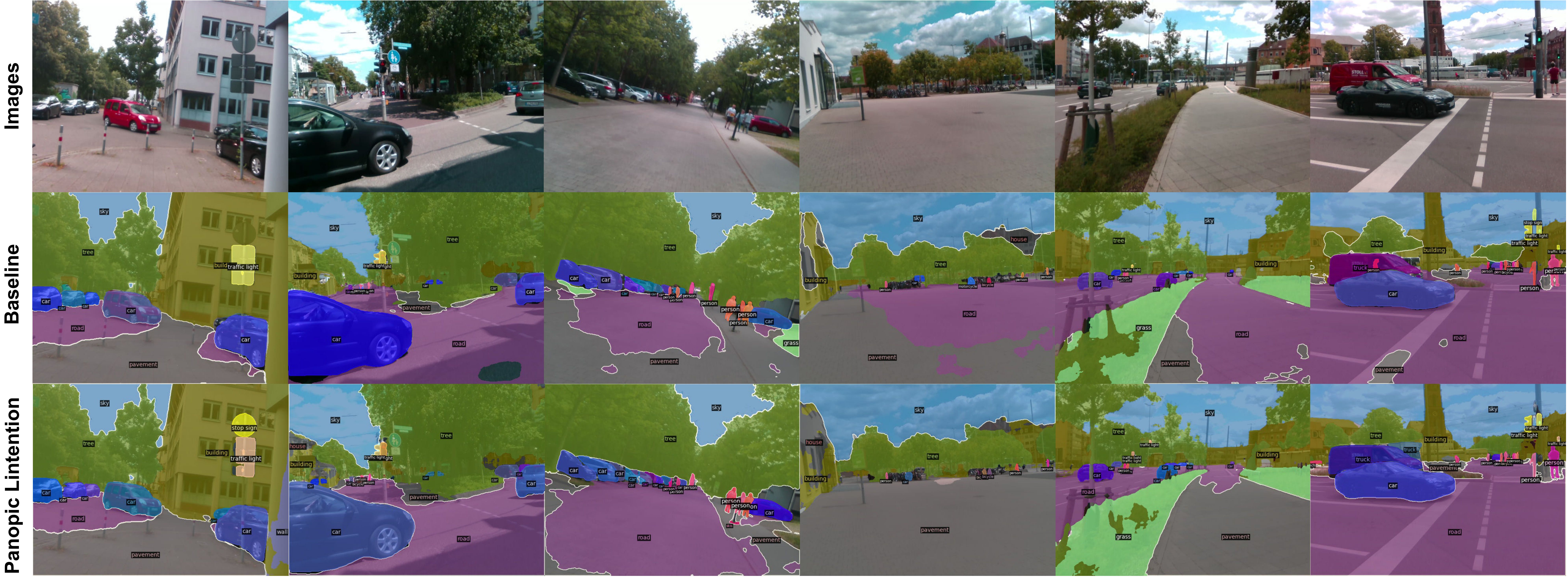}
    \vskip-1ex
    \caption{Comparison with the baseline Panoptic FPN on real-world images.
    Raw RGB image, baseline results, 
    and Panoptic Lintention results are organized from top to bottom.
    Best viewed in color. 
    }
    \label{fig:real_world_cmp}
    \vskip-4ex
\end{figure*}
As our main contributions lie in the semantic branch,
we  analyze the complexity of the baseline semantic branch and our proposed four semantic branches, which does not depend on the content of the specific image. 
We
feed an image of a fixed resolution to study the complexity
of the semantic branch only.
As indicated in Table~\ref{tab:flops_params_sem_head}, all of the four designed semantic branches are significantly more efficient than the baseline counterpart in terms of GFLOPs and the number of parameters.
Surprisingly, the Lintention Semantic branch has $25 \%$ fewer parameters than the baseline and requires $10 \%$ fewer GFLOPs for a $960 \times 1280$ image. 
Therefore, PyConv, VerConv, VerConSep, most notably the Lintention can serve as a more powerful substitute for the standard convolution without increasing the time complexity and model complexity.

\subsection{Real-World Test}
We design a compact wearable assistive system  as illustrated in the Fig.~\ref{fig:wearable_system}(a).

We invite a blindfolded person wearing the assistive system to navigate around the campus.
Meanwhile, the light-weight yet powerful RGB-D camera~\cite{yang2018unifying} captures a wide variety of scenes in the outdoor environment from the perspective of visually impaired people and  generates 30 RGB images at the resolution of 480×640.
The images captured by the RGB-D camera are processed by an NVIDIA Jetson AGX Xavier Developer Kit with a powerful CPU (8-core ARM v8.2 64-bit CPU) and a GPU (512-core Volta GPU with Tensor Cores).
the panoptic perception results are communicated to visually impaired people in the form of acoustic feedback~\cite{hu2020comparative} through the bone-conduction earphones integrated into the glasses. 
Our proposed framework acts as the first attempt to cover navigational perception needs for visually impaired people by using panoptic segmentation.

Various segmentation results for the images collected by the blindfolded person  are visualized in Fig.~\ref{fig:real_world_cmp}, 
which evidently showcase that the panoptic segmentation overcomes the fundamental flaws of
both instance and semantic segmentation by means of outputting a comprehensive pixel-level set of interpretations.
Therefore, the panoptic segmentation serves as an ideal tool for helping the visually impaired to push their environmental perception to a higher level.
As depicted in the Fig.~\ref{fig:real_world_cmp},
the safety-critical objects like cars and pedestrians are well recognized by baseline and our proposed models,
whereas only our proposed Panoptic Lintention Net can provide a reasonably satisfactory segmentation on the pavement and road which are essential for visually impaired people to safely participate in the traffic like a sighted person.

\section{Conclusion}
In this paper, we attempt to utilize panoptic segmentation to navigate visually impaired people by offering both things and stuff awareness in their proximity efficiently.
To this end,  we propose the Lintention which can learn long-range interactions in linear time using linear space.
Based on the Lintention, we devise the Panoptic Lintention Network for efficiently covering the navigational perception needs of the visually impaired by panoptic segmentation via our proposed wearable assistive system.
Experiments on the COCO dataset and
the real-world test 
indicate that our system based on the Panoptic Lintention Net accomplishes a relatively stable and exceptionally remarkable panoptic segmentation for visually impaired efficiently.

\bibliographystyle{IEEEtran}
\bibliography{bib.bib}

\begin{thebibliography}{10}
\providecommand{\url}[1]{#1}
\csname url@samestyle\endcsname
\providecommand{\newblock}{\relax}
\providecommand{\bibinfo}[2]{#2}
\providecommand{\BIBentrySTDinterwordspacing}{\spaceskip=0pt\relax}
\providecommand{\BIBentryALTinterwordstretchfactor}{4}
\providecommand{\BIBentryALTinterwordspacing}{\spaceskip=\fontdimen2\font plus
\BIBentryALTinterwordstretchfactor\fontdimen3\font minus
  \fontdimen4\font\relax}
\providecommand{\BIBforeignlanguage}[2]{{%
\expandafter\ifx\csname l@#1\endcsname\relax
\typeout{** WARNING: IEEEtran.bst: No hyphenation pattern has been}%
\typeout{** loaded for the language `#1'. Using the pattern for}%
\typeout{** the default language instead.}%
\else
\language=\csname l@#1\endcsname
\fi
#2}}
\providecommand{\BIBdecl}{\relax}
\BIBdecl

\bibitem{bourne2017magnitude}
R.~R. Bourne \emph{et~al.}, ``Magnitude, temporal trends, and projections of
  the global prevalence of blindness and distance and near vision impairment: a
  systematic review and meta-analysis,'' \emph{The Lancet Global Health}, 2017.

\bibitem{yang2018unifying}
K.~Yang \emph{et~al.}, ``Unifying terrain awareness for the visually impaired
  through real-time semantic segmentation,'' \emph{Sensors}, 2018.

\bibitem{rodriguez2012assisting}
A.~Rodr{\'\i}guez, J.~J. Yebes, P.~F. Alcantarilla, L.~M. Bergasa,
  J.~Almaz{\'a}n, and A.~Cela, ``Assisting the visually impaired: obstacle
  detection and warning system by acoustic feedback,'' \emph{Sensors}, 2012.

\bibitem{schwarze2015intuitive}
T.~Schwarze, M.~Lauer, M.~Schwaab, M.~Romanovas, S.~B{\"o}hm, and
  T.~J{\"u}rgensohn, ``An intuitive mobility aid for visually impaired people
  based on stereo vision,'' in \emph{ICCVW}, 2015.

\bibitem{watson2020footprints}
J.~Watson, M.~Firman, A.~Monszpart, and G.~J. Brostow, ``Footprints and free
  space from a single color image,'' in \emph{CVPR}, 2020.

\bibitem{lee2020pedestrian}
K.~Lee, D.~Sato, S.~Asakawa, H.~Kacorri, and C.~Asakawa, ``Pedestrian detection
  with wearable cameras for the blind: A two-way perspective,'' in \emph{CHI},
  2020.

\bibitem{he2017mask}
K.~He, G.~Gkioxari, P.~Doll{\'a}r, and R.~Girshick, ``Mask r-cnn,'' in
  \emph{ICCV}, 2017.

\bibitem{long2019unifying}
N.~Long, K.~Wang, R.~Cheng, W.~Hu, and K.~Yang, ``Unifying obstacle detection,
  recognition, and fusion based on millimeter wave radar and rgb-depth sensors
  for the visually impaired,'' \emph{RSI}, 2019.

\bibitem{kirillov2019panoptic}
A.~Kirillov, K.~He, R.~Girshick, C.~Rother, and P.~Doll{\'a}r, ``Panoptic
  segmentation,'' in \emph{CVPR}, 2019.

\bibitem{cao2020rapid}
Z.~Cao, X.~Xu, B.~Hu, and M.~Zhou, ``Rapid detection of blind roads and
  crosswalks by using a lightweight semantic segmentation network,''
  \emph{T-ITS}, 2020.

\bibitem{duh2020v}
P.-J. Duh, Y.-C. Sung, L.-Y.~F. Chiang, Y.-J. Chang, and K.-W. Chen, ``V-eye: A
  vision-based navigation system for the visually impaired,'' \emph{TMM}, 2020.

\bibitem{yohannes2019content}
E.~Yohannes, T.~K. Shih, and C.-Y. Lin, ``Content-aware video analysis to guide
  visually impaired walking on the street,'' in \emph{IVIC}, 2019.

\bibitem{yang2018intersection}
K.~Yang, R.~Cheng, L.~M. Bergasa, E.~Romera, K.~Wang, and N.~Long,
  ``Intersection perception through real-time semantic segmentation to assist
  navigation of visually impaired pedestrians,'' in \emph{ROBIO}, 2018.

\bibitem{kirillov2019pfpn}
A.~Kirillov, R.~Girshick, K.~He, and P.~Doll{\'a}r, ``Panoptic feature pyramid
  networks,'' in \emph{CVPR}, 2019.

\bibitem{li2019attention}
Y.~Li \emph{et~al.}, ``Attention-guided unified network for panoptic
  segmentation,'' in \emph{CVPR}, 2019.

\bibitem{chen2020banet}
Y.~Chen \emph{et~al.}, ``Banet: Bidirectional aggregation network with
  occlusion handling for panoptic segmentation,'' in \emph{CVPR}, 2020.

\bibitem{cheng2020panoptic}
B.~Cheng \emph{et~al.}, ``Panoptic-deeplab: A simple, strong, and fast baseline
  for bottom-up panoptic segmentation,'' in \emph{CVPR}, 2020.

\bibitem{vaswani2017attention}
A.~Vaswani \emph{et~al.}, ``Attention is all you need,'' \emph{NeurIPS}, 2017.

\bibitem{laue2020simple}
S.~Laue, M.~Mitterreiter, and J.~Giesen, ``A simple and efficient tensor
  calculus,'' in \emph{AAAI}, 2020.

\bibitem{lin2014microsoft}
T.-Y. Lin \emph{et~al.}, ``Microsoft coco: Common objects in context,'' in
  \emph{ECCV}, 2014.

\bibitem{duta2020pyramidal}
I.~C. Duta, L.~Liu, F.~Zhu, and L.~Shao, ``Pyramidal convolution: rethinking
  convolutional neural networks for visual recognition,''
  \emph{arXiv:2006.11538}, 2020.

\bibitem{hu2018squeeze}
J.~Hu, L.~Shen, and G.~Sun, ``Squeeze-and-excitation networks,'' in
  \emph{CVPR}.\hskip 1em plus 0.5em minus 0.4em\relax IEEE, 2018.

\bibitem{he2016deep}
K.~He, X.~Zhang, S.~Ren, and J.~Sun, ``Deep residual learning for image
  recognition,'' in \emph{CVPR}, 2016.

\bibitem{carion2020end}
N.~Carion, F.~Massa, G.~Synnaeve, N.~Usunier, A.~Kirillov, and S.~Zagoruyko,
  ``End-to-end object detection with transformers,'' in \emph{ECCV}, 2020.

\bibitem{hu2020comparative}
W.~Hu \emph{et~al.}, ``A comparative study in real-time scene sonification for
  visually impaired people,'' \emph{Sensors}, 2020.

\end{thebibliography}

\end{document}